\documentclass[11pt,a4paper]{article}
\usepackage[hyperref]{acl2017}
\usepackage{times}
\usepackage{latexsym}

\usepackage{url}

\aclfinalcopy 

\usepackage{url}
\usepackage{todonotes}

\title{FEUP at SemEval-2017 Task 5: Predicting Sentiment Polarity and Intensity with Financial Word Embeddings}

\author{Pedro Saleiro$^{1,2}$, Eduarda Mendes Rodrigues$^{1}$, Carlos Soares$^{1,3}$, Eug\'enio Oliveira$^{1,2}$ \\
  Faculdade de Engenharia da Universidade do Porto$^1$, \\ LIACC$^2$, INESC-TEC$^3$\\
  Rua Dr. Roberto Frias, s/n, Porto, Portugal \\
  {\tt \{pssc,eduarda,csoares,eco\}@fe.up.pt} \\
}  

\date{}

\begin{document}
\maketitle
\begin{abstract}

This paper presents the approach developed at the Faculty of Engineering of University of Porto, to participate in SemEval 2017, Task 5: Fine-grained Sentiment Analysis on Financial Microblogs and News. 
The task consisted in predicting a real continuous variable from -1.0 to +1.0 representing the polarity and intensity of sentiment concerning companies/stocks mentioned in short texts. We modeled the task as a regression analysis problem and combined traditional techniques such as pre-processing short texts, bag-of-words representations and lexical-based features with enhanced financial specific bag-of-embeddings. We used an external collection of tweets and news headlines mentioning companies/stocks from S\&P 500 to create financial word embeddings which are able to capture domain-specific syntactic and semantic similarities. The resulting approach obtained a cosine similarity score of 0.69 in sub-task 5.1 - Microblogs and 0.68 in sub-task 5.2 - News Headlines.

\end{abstract}

\section{Introduction}

Sentiment Analysis on financial texts has received increased attention in recent years~\citep{nardo2016}. Neverthless, there are some challenges yet to overcome~\citep{smailovic2014stream}. Financial texts, such as microblogs or newswire, usually contain highly technical and specific vocabulary or jargon, making the develop of specific lexical and machine learning approaches necessary. Most of the research in Sentiment Analysis in the financial domain has focused in analyzing subjective text, labeled with explicitly expressed sentiment.

However, it is also common to express financial sentiment in an implicit way. Business news stories often refer to events that  might indicate a positive or negative impact, such as in the news title ``company X will cut 1000 jobs''. Economic indicators, such as unemployment and future state modifiers such as drop or increase can also provide clues on the implicit sentiment~\citep{musat2010impact}. Contrary to explicit expressions (subjective utterances), factual text types often contain objective statements that convey a desirable or undesirable fact~\citep{liu2012sentiment}.

Recent work proposes to consider all types of implicit sentiment expressions ~\citep{van2015fine}.
The authors created a fine grained sentiment annotation procedure to identify polar expressions (implicit and explicit expressions of positive and negative sentiment). A target (company of interest) is identified in each polar expression to identify the sentiment expressions that are relevant. The annotation procedure also collected information about the polarity and the intensity of the sentiment expressed towards the target. However, there is still no automatic approach, either lexical-based or machine learning based, that tries to model this annotation scheme. 

\begin{table*}[t]
\centering
\begin{tabular}{|l|l|l|l|}
\hline
  Sub-task & Company & Text Span & Sentiment Score\\
  \hline
  5.1 - Microblogs & JPMorgan &``its time to sell banks" & -0.763 \\
  5.2 - Headlines & Glencore &``Glencore's annual results beat forecasts" & +0.900 \\
  \hline
\end{tabular}
\caption{Training set examples for both sub-tasks.} \label{examp}
\end{table*}

In this work, we propose to tackle the aforementioned problem by taking advantage of unsupervised learning of word embeddings in financial tweets and financial news headlines to construct a domain-specific syntactic and semantic representation of words. We combine bag-of-embeddings with traditional approaches, such as pre-processing techniques, bag-of-words and financial lexical-based features to train a regressor for sentiment polarity and intensity. We study how different regression algorithms perform using all features in two different sub-tasks at SemEval-2017 Task 5: microblogs and news headlines mentioning companies/stocks. Moreover, we compare how different combinations of features perform in both sub-tasks. The system source code and word embeddings developed for the competition are publicly available.%
\footnote{\url{https://github.com/saleiro/Financial-Sentiment-Analysis}}

The remainder of the paper is organized as follows. We start by describing SemEval-2017 Task 5 and how we created financial-specific word embeddings. In Section \ref{app} we present the implementation details of the system created for the competition followed by the experimental setup. We then present the experimental results and analysis, ending with the conclusions of this work.

\section{Task Description}

The task 5 of SemEval 2017~\citep{semeval2017task5} consisted of fine-grained sentiment analysis of financial short texts and it was divided in two  sub-tasks based on the type of text. Sub-task 5.1 -- Microblogs -- consisted of stocktwits and tweets focusing on stock market events and assessments from investors and traders. Companies/stocks were identified using stock symbols, the so called cashtags, e.g.``\$AMZN'' for the company Amazon.com, Inc. Sub-task 5.2 -- News Headlines -- consisted of sentences extracted from Yahoo Finance and other financial news sources on the internet. In this case, companies/stocks were identified using their canonical name and were previously annotated by the task organizers.

The goal of both sub-tasks was the following: predict the sentiment polarity and intensity for each of the companies/stocks mentioned in a short text instance (microblog message or news sentence). The sentiment score is a real continuous variable in the range of -1.0 (very negative/bearish) to +1.0 (very positive/bullish), with 0.0 designating neutral sentiment. Table~\ref{examp} presents two examples from the training set. Task organizers provided 1700 microblog messages for training and 800 messages for testing in sub-task 5.1, while in sub-task 5.2, 1142 news sentences were provided for training and 491 for testing. Submissions were evaluated using the cosine similarity~\citep{semeval2017task5}. 

\section{Financial Word Embeddings}
\citeauthor{mikolov2013efficient} \shortcite{mikolov2013efficient} created word2vec, a computational efficient method to learn distributed representation of words, where each word is represented by a distribution of weights (embeddings) across a fixed set of dimensions. Furthermore,~\citeauthor{mikolov2013distributed} \shortcite{mikolov2013distributed} showed that this representation is able to encode syntactic and semantic similarities in the embedding space. 

The training objective of the skip-gram model, defined by ~\citeauthor{mikolov2013distributed} \shortcite{mikolov2013distributed}, is to learn the target word representation (embeddings) that maximize the prediction of its surrounding words in a context window. Given the $w_t$ word in a vocabulary the objective is to maximize the average log probability: 

\begin{equation}
\frac{1}{T}  \sum_{t=1}^{T}  \sum_{-c \leq j \leq  c, j \neq 0} \textnormal{log } P(w_{t+j} | w_t)
\end{equation}

where $c$ is the size of the context window, $T$ is the total number of words in the vocabulary and $w_{t+j}$ is a word in the context window of $w_t$. After training, a low dimensionality embedding matrix $\textbf{E}$ encapsulates information about each word in the vocabulary and its use (surrounding contexts).

We used word2vec to learn word embeddings in the context of financial texts using unlabeled tweets and news headlines mentioning companies/stocks from S\&P 500. Tweets were collected using the Twitter streaming API with cashtags of stocks titles serving as request parameters. Yahoo Finance API was used for requesting financial news feeds by querying the canonical name of companies/stocks. The datasets comprise a total of 1.7M tweets and 626K news titles.

We learned separate word embeddings for tweets and news headlines using the skip-gram model. We tried several configurations of word2vec hyperparameters. The setup resulting in the best performance in both sub-tasks was skip-gram with 50 dimensions, removing words occurring less than 5 times,
using a context window of 5 words and 25 negative samples per positive example. 

Even though the text collections for training embeddings were relatively small, the resulting embedding space exhibited the ability to capture semantic word similarities in the financial context. We performed simple algebraic operations to capture semantic relations between words, as described in ~\citeauthor{mikolov2013linguistic} \shortcite{mikolov2013linguistic}.  For instance, the skip-gram model trained on tweets shows that vector (``bearish'') - vector(``loss'') + vector(``gain'') results in vector (``bullish'') as most similar word representation. 

\section{Approach} \label{app}
In this section we describe the implementation details of the proposed approach.

\subsection{Pre-Processing}
A set of pre-processing operations are applied to every microblog message and news sentence in the training/test sets of sub-tasks 5.1 and 5.2, as well as in the external collections for training word embeddings:

\begin{itemize}
\item \textbf{Character encoding and stopwords}: every message and headline was encoded in UTF-8. Standard english stopword removal is also applied.

\item \textbf{Company/stock and cash obfuscation}: both cashtags and canonical company names strings were replaced by the string \textit{\_company\_}. Dollar or Euro signs followed by numbers were replaced by the string \textit{\_cash\_amount\_}.

\item \textbf{Mapping numbers and signs}: numbers were mapped to strings using bins (0-10, 10-20, 20-50, 50-100, $>$100). Minus and plus signs were coverted to \textit{minus} and \textit{plus}, ``B'' and ``M'' to \textit{billions} and \textit{millions}, respectively. The \% symbol was converted to \textit{percent}. Question and exclamation marks were also converted to strings.

\item \textbf{Tokenization, punctuation, lowercasing}: tokenization was performed using Twokenizer~\citep{gimpel2011part}, the remaining punctuation was removed and all characters were converted to lowercase.
\end{itemize}

\subsection{Features}

We combined three different group of features: bag-of-words, lexical-based features and bag-of-embeddings. 

\begin{itemize}
\item \textbf{Bag-of-words}: we apply standard bag-of-words as features. We tried unigrams, bi-grams and tri-grams with unigrams proving to obtain higher cosine similarity in both sub-tasks.

\item \textbf{Sentiment lexicon features}: we incorporate knowledge from manually curated sentiment lexicons for generic Sentiment Analysis as well as lexicons tailored for the financial domain. The Laughran-Mcdonald financial sentiment dictionary~\cite{bodnaruk2015using} has several types of word classes: positive, negative, constraining, litigious, uncertain and modal. For each word class we create a binary feature for the match with a word in a microblog/headline and a polarity score feature (positive - negative normalized by the text span length).  As a general-purpose sentiment lexicon we use MPQA~\citep{wilson2005recognizing} and created binary features for positive, negative and neutral words, as well as, the polarity score feature.

\item \textbf{Bag-of-Embeddings}: we create bag-of-embeddings by taking the average of word vectors for each word in a text span. We used the corresponding embedding matrix trained on external Twitter and Yahoo Finance collections for sub-task 5.1 and sub-task 5.2, respectively. 

\end{itemize}

\section{Experimental Setup}

In order to avoid overfitting we created a validation set from the original training datasets provided by the organizers. We used a 80\%-20\% split and sampled the validation set using the same distribution as the original training set.
We sorted the examples in the training set by the target variable values and skipped every 5 examples. Results are evaluated using Cosine similarity~\citep{semeval2017task5} and Mean Average Error (MAE). The former gives more importance to differences in the polarity of the predicted sentiment while the latter is concerned with how well the system predicts the intensity of the sentiment.


We opted to model both sub-tasks as single regression problems. Three different regressors were applied: Random Forests (RF), Support Vector Machines (SVM) and MultiLayer Perceptron (MLP). Parameter tuning was carried using 10 fold cross validation on the training sets.

\section{Results and Analysis}

In this section we present the experimental results obtained in both sub-tasks. We provide comparison of different learning algorithms using all features, as well as, a comparison of different subsets of features, to understand the information contained in each of them and also how they complement each other.

\subsection{Task 5.1 - Microblogs}
\label{sec:length}

Table~\ref{res1} presents the results obtained using all features in both validation set and test sets.  Results in the test set are worse than in the validation set with the exception to MLP. The official score obtained in sub-task 5.1 was 0.6948 using Random Forests (RF), which is the regressor that achieves higher cosine similarity and lower MAE in both training and validation set.

\begin{table}[h]
\centering
\begin{tabular}{|l|l|l|l|}
\hline
  Regressor & Set & Cosine & MAE\\
  \hline
  RF & Val & 0.7960 & 0.1483 \\
  RF & \textbf{Test} & \textbf{0.6948} & \textbf{0.1886} \\
  SVR & Val & 0.7147 & 0.1944 \\
  SVR & \textbf{Test} & 0.6227 & 0.2526 \\
  MLP & Val & 0.6720 & 0.2370 \\
  MLP & \textbf{Test} & 0.6789 & 0.2132 \\
  \hline
\end{tabular}
\caption{Microblog results with all features on validation and test sets.} \label{res1}
\end{table}

We compared the results obtained with different subsets of features using the best regressor, RF, as depicted in Table~\ref{res2}. Interestingly, bag-of-words (BoW) and bag-of-embeddings (BoE) complement each other, obtaining better cosine similarity than the system using all features. Financial word embeddings (BoE) capture relevant information regarding the target variables. As a single group of features it achieves a cosine similarity of 0.6118 and MAE of 0.2322. It is also able to boost the overall performance of BoW with gains of more than 0.06 in cosine similarity and reducing MAE more than 0.03. 

The individual group of features with best performance is Bag-of-words while the worst is a system trained using Lex (only lexical-based features). While Lex alone exhibits poor performance, having some value but marginal, when combined with another group of features, it improves the results of the latter, as in the case of BoE + Lex and BoW + Lex.

\begin{table}[h]
\centering
\begin{tabular}{|l|l|l|}
\hline
  Features &  Cosine & MAE\\
  \hline
  Lex &  0.3156 & 0.3712 \\
  BoE &  0.6118 & 0.2322 \\
  BoW &  0.6386 &0.2175 \\
  BoE + Lex &  0.6454 & 0.2210 \\
  Bow + Lex &  0.6618 & 0.2019 \\
  Bow + BoE &  \textbf{0.7023} & \textbf{0.1902} \\
  All &  0.6948 & 0.1886 \\

  \hline
\end{tabular}
\caption{Features performance breakdown on test set using RF.} \label{res2}
\end{table}

\subsection{Task 5.2 - News Headlines}

Results obtained in news headlines are very different from the ones of the previous sub-task, proving that predicting sentiment polarity and intensity in news headlines is a complete different problem compared to microblogs.
Table~\ref{res3} shows that MLP obtains the best results in the test set using both metrics while SVR obtains the best performance in the validation set. The best regressor of sub-task 5.1, RF is outperformed by both SVR and MLP. The official result obtained at sub-task 5.2 was a cosine similarity of 0.68 using MLP. 

\begin{table}[h]
\centering
\begin{tabular}{|l|l|l|l|}
\hline
  Regressor & Set & Cosine & MAE\\
  \hline
  RF & Val & 0.5316 &0.2539 \\
  RF & \textbf{Test} & 0.6562 & 0.2258\\
  SVR & Val & 0.6397 &0.2422 \\
  SVR & \textbf{Test} &  0.6621 & 0.2424 \\
  MLP & Val &  0.6176 &0.2398 \\
  MLP & \textbf{Test} & \textbf{0.6800} & \textbf{0.2271} \\

  \hline
\end{tabular}
\caption{News Headlines results with all features on validation and test sets.} \label{res3}
\end{table}

Table~\ref{res4} shows the results of the different groups of features in sub-task 5.2 for MLP regressor. The most evident observation is that word embeddings are not effective in this scenario. On the other hand, lexical based features have significantly better performance in news headlines than in microblogs. Despite this, the best results are obtained using all features.

\begin{table}[h]
\centering
\begin{tabular}{|l|l|l|}
\hline
  Features &  Cosine & MAE\\
  \hline
  BoE &  0.0383 & 0.3537 \\
  Lex &  0.5538 & 0.2788 \\
  BoW &  0.6420 &0.2364 \\
  BoE + Lex &  0.5495 & 0.2830 \\
  BoW + Lex &  \textbf{0.6733} & \textbf{0.2269} \\
  BoW + BoE &   0.6417 &  0.2389 \\
  All & 0.6800 & 0.2271 \\

  \hline
\end{tabular}
\caption{Features performance breakdown on test set using MLP.} \label{res4}
\end{table}

\subsection{Analysis}

Financial word embeddings were able to encapsulate valuable information in sub-task 5.1 - Microblogs but not so much in the case of sub-task 5.2 - News Headlines. We hypothesize that as we had access to a much smaller dataset ($\sim 600$K) for training financial word embeddings for news headlines, this resulted in reduced ability to capture semantic similarities in the financial domain. Other related works in Sentiment Analysis usually take advantage of a much larger dataset for training word embeddings \citep{deriu2016swisscheese}.

On the other hand, lexical features showed poor performance in microblog texts but seem to be very useful using news headlines. The fact that microblogs have poor grammatically constructed texts, slang and informal language reveals that financial lexicals created using well written and formal financial reports, result better in news headlines rather than in microblog texts.

\section{Conclusions}

Work reported in this paper is concerned with the problem of predicting sentiment polarity and intensity of financial short texts. Previous work showed that sentiment is often depicted in an implicit way in this domain. We created financial-specific continuous word representations in order to obtain domain specific syntactic and semantic relations between words. We combined traditional bag-of-words and lexical-based features with bag-of-embeddings to train a regressor of both sentiment polarity and intensity. Results show that different combination of features attained different performances on each sub-task. Future work will consist on collecting larger external datasets for training financial word embeddings of both microblogs and news headlines. We also have planned to perform the regression analysis using Deep Neural Networks.

\bibliography{semeval2017}
\bibliographystyle{acl_natbib}

\end{document}